\documentclass[conference]{IEEEtran}

\ifCLASSOPTIONcompsoc
  \usepackage[nocompress]{cite}
\else
  \usepackage{cite}
\fi

\ifCLASSINFOpdf
   \usepackage[pdftex]{graphicx}
\else
\fi

\usepackage{array}
\usepackage{xcolor}
\usepackage{colortbl}

\ifCLASSOPTIONcompsoc
 \usepackage[caption=false,font=footnotesize,labelfont=sf,textfont=sf]{subfig}
\else
 \usepackage[caption=false,font=footnotesize]{subfig}
\fi

\usepackage{booktabs}
\usepackage{url}
\usepackage{siunitx}
\usepackage{enumitem}
\usepackage{etoolbox}
\usepackage{booktabs}
\usepackage{makecell}
\usepackage{amsmath}
\usepackage{placeins}
\usepackage{graphicx}
\usepackage{float} 
\usepackage{textcomp}
\usepackage{comment}
\usepackage{algorithm}
\usepackage{algpseudocode} % This loads algorithmicx
\usepackage{float}
\usepackage{placeins}
\usepackage{makecell}
\usepackage[normalem]{ulem} % For \sout
\usepackage{multirow}
\definecolor{darkgreen}{rgb}{0.0, 0.5, 0.0}
\definecolor{robo_blue}{RGB}{66, 133, 244}
\definecolor{robo_red}{RGB}{231, 66, 52}
\definecolor{robo_yellow}{RGB}{251, 189, 5}
\definecolor{robo_green}{RGB}{51, 168, 82}
\definecolor{robo_gray}{RGB}{165, 165, 165}

\usepackage[T1]{fontenc}
\usepackage{amsmath,amssymb,amsfonts} % amssymb might be needed if you add symbols later
\usepackage{siunitx}
\usepackage{caption} % Needed for caption within table*
\usepackage{makecell}
\usepackage{url}        % If using URLs
\usepackage{microtype}  % Good for justification
\usepackage[pagebackref=false,breaklinks=true,colorlinks=true,bookmarks=false]{hyperref} % Load hyperref near the end

\usepackage{titlesec}
\titleformat{\section}
  {\normalfont\fontsize{11}{12}\bfseries}{\thesection}{1em}{}

\usepackage[pagebackref=false,breaklinks=true,colorlinks=true,bookmarks=false]{hyperref}
\hypersetup{
    colorlinks=true,
    linkcolor=blue,
    filecolor=magenta,
    urlcolor=cyan,
    citecolor=green, % Changed cite color for visibility
    pdftitle={Pruning},
    pdfpagemode=FullScreen,
    }

\begin{document}

\title{Teacher-Guided One-Shot Pruning via Context-Aware Knowledge Distillation}

% \author{Md. Samiul Alim, Sharjil Khan, Amrijit Biswas, Shafin Rahman, Fuad Rahman, Nabeel Mohammed}
\author{
Md. Samiul Alim$^1$ \qquad Sharjil Khan$^1$ \qquad Amrijit Biswas$^1$ \\
Fuad Rahman$^2$ \qquad Shafin Rahman$^1$ \qquad Nabeel Mohammed$^1$ \\\\
$^1$ Apurba-NSU R\&D Lab, \\
Department of Electrical and Computer Engineering, \\
North South University, Dhaka, Bangladesh \\\\
$^2$ Apurba Technologies, Sunnyvale, CA 94085, USA \\\\
{\tt\small \{samiul.alim01, sharjil.khan, amrijit.biswas01, shafin.rahman, nabeel.mohammed\}@northsouth.edu} \\
{\tt\small fuad@apurbatech.com}
}

% --- End Author Information ---

\IEEEtitleabstractindextext{%
\begin{abstract}
Unstructured pruning remains a powerful strategy for compressing deep neural networks, yet it often demands iterative train--prune--retrain cycles, resulting in significant computational overhead. To address this challenge, we introduce a novel teacher-guided pruning framework that tightly integrates Knowledge Distillation (KD) with importance score estimation. Unlike prior approaches that apply KD as a post-pruning recovery step, our method leverages gradient signals informed by the teacher during importance score calculation to identify and retain parameters most critical for both task performance and knowledge transfer. Our method facilitates a one-shot global pruning strategy that efficiently eliminates redundant weights while preserving essential representations. After pruning, we employ sparsity-aware retraining with and without KD to recover accuracy without reactivating pruned connections. Comprehensive experiments across multiple image classification benchmarks, including CIFAR-10, CIFAR-100, and TinyImageNet, demonstrate that our method consistently achieves high sparsity levels with minimal performance degradation. Notably, our approach consistently outperforms state-of-the-art baselines such as EPG, and EPSD at high sparsity levels, while offering a more computationally efficient alternative to iterative pruning schemes like COLT. The proposed framework offers a computation-efficient, performance-preserving solution well-suited for deployment in resource-constrained environments.
\end{abstract}
\begin{IEEEkeywords}
 Pruning, Knowledge Distillation (KD), One-Shot Pruning, Lottery Ticket Hypothesis (LTH)
\end{IEEEkeywords}

}

\maketitle
\thispagestyle{plain} % Ensure first page is plain
\pagestyle{plain}    % Ensure subsequent pages are plain

% For peer review papers, this IEEEtran command inserts a page break and
\IEEEdisplaynontitleabstractindextext
\IEEEpeerreviewmaketitle

\section{Introduction}
\label{sec:introduction}
\begin{figure}[h]
    \centering
    % First figure
    \begin{minipage}{0.48\textwidth}
        \centering
        \includegraphics[width=\textwidth]{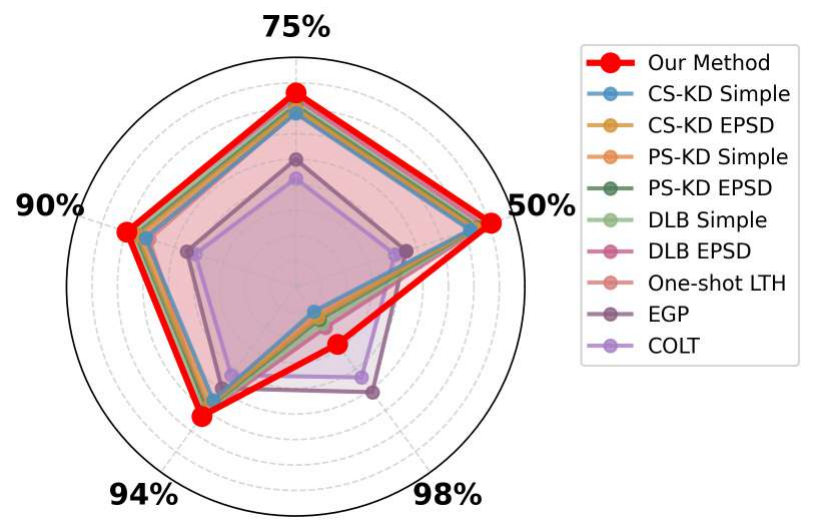}
\caption{Overview of accuracy comparison between our method and existing approaches across sparsity levels (50\%, 75\%, 90\%, 94\%, 98\%). In the radar plot, a larger enclosed area indicates higher accuracy, where \textbf{\textcolor{red}{Our Method}} consistently achieves superior performance on CIFAR-10 datasets, clearly outperforming competing baselines. Detailed results including CIFAR-10 and other datasets are presented in the Results section.}

        \label{fig:accuracy_comparison}
    \end{minipage}
\end{figure}

Deep Neural Networks (DNNs), particularly Convolutional Neural Networks (CNNs), have achieved state-of-the-art performance across numerous computer vision tasks such as image classification~\cite{iandola2016squeezenet, lecun2015deep}, object detection, and semantic segmentation~\cite{chen2017deeplab}. However, these performance gains often come with a high computational and memory cost~\cite{lecun2015deep}, limiting their deployment on resource-constrained environments such as mobile devices and edge platforms. To address this, the community has explored various model compression techniques aimed at reducing model complexity while retaining performance. Common approaches include parameter quantization~\cite{gholami2022survey}, efficient architecture design~\cite{iandola2016squeezenet}, knowledge distillation (KD)~\cite{hinton2015distilling}, and network pruning~\cite{han2015deep_compression, blalock2020state}.

Among these, network pruning stands out for its ability to significantly reduce parameter count by removing redundant weights, filters, or neurons. It is broadly categorized into structured pruning~\cite{li2016pruning, luo2017entropy}, which removes entire channels or filters for hardware efficiency, and unstructured pruning~\cite{han2015learning, frankle2018lottery}, which eliminates individual weights and can achieve higher sparsity levels. While structured pruning improves practical deployment, it often sacrifices fine-grained control and performance. Unstructured pruning, in contrast, allows finer sparsity control and higher compression ratios but typically relies on less informative heuristics such as magnitude-based criteria~\cite{han2015learning}{}-to determine parameter importance.

Several limitations persist in the current landscape of unstructured pruning. First, importance estimation is often based on heuristics (e.g., weight magnitude) that fail to capture dynamic learning signals. This limitation is evident in approaches like the Lottery Ticket Hypothesis (LTH)~\cite{frankle2018lottery}, which require multiple train-prune-retrain iterations to isolate performant subnetworks, resulting in excessive computational cost. Second, while gradient-based criteria~\cite{lecun1990optimal, molchanov2017pruning} offer a more principled alternative by leveraging sensitivity, they often rely on noisy single-step gradients and are detached from any auxiliary supervision such as that from a teacher network. Third, knowledge distillation (KD) - a powerful framework to transfer knowledge from a large teacher to a smaller student - has been predominantly used as a post-pruning recovery tool~\cite{polino2018model, aghli2021combining}, rather than being integrated into the pruning process itself. Consequently, pruning decisions are made without the benefit of the teacher’s informative soft targets. While EPSD~\cite{chen2024epsdearlypruningselfdistillation} addresses this challenge through self-distillation-aware early pruning, our approach extends this by using a pretrained teacher to actively guide the pruning process via task-aligned gradients.

\textbf{Our goal is to address these limitations} by introducing a teacher-guided unstructured pruning framework that leverages both gradient sensitivity and knowledge distillation in a unified pipeline. Unlike iterative methods such as LTH~\cite{frankle2018lottery} and COLT~\cite{hossain2022colt}, which require costly train-prune-retrain cycles, our approach enables efficient one-shot global pruning. While prior efforts like EPSD~\cite{chen2024epsdearlypruningselfdistillation} incorporate self-distillation to improve early-stage pruning, they lack explicit teacher supervision during importance estimation. In contrast, we incorporate the teacher's guidance directly into the pruning signal through a distillation-aware loss function specifically, Context-Aware Kullback Leibler Divergence (CA-KLD)~\cite{liu2023context} augmented with logit normalization~\cite{sun2024logit} for stable optimization. Gradient signals informed by this combined loss are aggregated using an exponential moving average with bias correction and used to rank parameter importance. The resulting scores enable aggressive one-shot pruning, followed by sparsity-aware retraining either with or without KD while strictly preserving the pruned structure.

The main contributions of our work are:
\begin{itemize}
    \item A novel teacher-guided gradient importance metric that utilizes gradients derived from both task loss and an advanced KD loss (CA-KLD).
    \item A demonstration of integrating advanced KD not just for post-pruning recovery, but as an active component guiding the identification of critical parameters during the importance score calculation.
    \item Our One-Shot Global Pruning method minimizes computational cost compared to iterative pruning methods such as Lottery Ticket Hypothesis~\cite{frankle2018lottery} and COLT~\cite{hossain2022colt}. Furthermore, our method outperforms six different methods evaluated in the EPSD~\cite{chen2024epsdearlypruningselfdistillation}, including CS-KD Simple, CS-KD EPSD, PS-KD Simple,PS-KD EPSD, DLB Simple, and DLB EPSD, as well as the EGP method~\cite{liao2023can}.
    \item Extensive empirical evaluation on benchmark datasets (CIFAR-10, CIFAR-100, TinyImageNet) using standard architectures (ResNet18, ResNet34), demonstrating competitive performance compared to other methods.
\end{itemize}
\section{Related Work}
\label{sec:related_work}

Deep neural networks, particularly convolutional neural networks (CNNs), have achieved remarkable success across various domains, but their growing computational and memory demands pose challenges for deployment on resource-constrained devices \cite{lecun2015deep}. This has motivated extensive research into model compression techniques, including parameter quantization \cite{gholami2022survey}, knowledge distillation (KD) \cite{hinton2015distilling}, designing compact architectures \cite{iandola2016squeezenet}, and network pruning \cite{han2015deep_compression, blalock2020state}. Our work focuses on pruning enhanced by KD.

\textbf{Network Pruning:} Network pruning reduces model complexity by removing redundant weights, neurons, or filters while minimizing performance loss. \textbf{Unstructured pruning} eliminates individual weights, often based on magnitude \cite{han2015learning, frankle2018lottery}, resulting in sparse networks that require specialized hardware or libraries for efficient inference. The Lottery Ticket Hypothesis (LTH) \cite{frankle2018lottery} suggests that dense networks contain sparse "winning ticket" subnetworks that can match the original model’s performance when trained from early initialization. \textbf{Structured pruning}, on the other hand, removes entire filters, channels, or layers \cite{li2016pruning, luo2017entropy}, producing smaller dense networks compatible with standard hardware, though potentially more detrimental to accuracy, especially in architectures like ResNets \cite{aghli2021combining}.  

Pruning criteria include magnitude-based methods \cite{han2015learning, frankle2018lottery}, activation-based metrics such as APoZ \cite{hu2016network_trimming} or channel entropy \cite{luo2017entropy, liao2023can}, and gradient-based approaches \cite{lecun1990optimal, molchanov2017pruning}. Our method belongs to the gradient-based category but incorporates teacher-student supervision, leveraging first-order gradients weighted by a task-distillation loss to estimate importance efficiently.

\textbf{Knowledge Distillation for Compression:} KD \cite{hinton2015distilling} transfers knowledge from a large teacher model to a smaller student using softened output probabilities, capturing richer inter-class relationships than hard labels alone. KD can improve the performance of compact models \cite{romero2015fitnets} and is often combined with pruning to help recover accuracy lost during compression \cite{polino2018model, aghli2021combining}. Advanced variants, such as Context Aware KLD (CA-KLD) \cite{liu2023context}, consider both teacher confidence and uncertainty to produce better-calibrated students.

\textbf{Our Approach in Context:}
Unlike conventional pruning strategies based on magnitude, activation sparsity, or entropy \cite{han2015deep_compression, hu2016network_trimming, luo2017entropy}, our framework embeds teacher-guided supervision directly into gradient-based importance estimation. By using smoothed first-order gradients aligned with CA-KLD soft targets \cite{liu2023context, sun2024logit}, we perform global one-shot pruning, reducing computational cost compared to iterative methods such as LTH \cite{frankle2018lottery} or COLT \cite{hossain2022colt}. Our dual retraining regimes standard and KD-aware—further ensure that pruned models retain high accuracy while preserving the benefits of compression.

\section{Methodology}
\label{sec:pipeline}

\subsection{Problem Formulation}
Let $\mathcal{D} = \{(x_i, y_i)\}_{i=1}^{N}$ denote a labeled training dataset, 
where $x_i \in \mathbb{R}^{h \times w \times c}$ are input images and 
$y_i \in \{1, \dots, C\}$ are the corresponding class labels. 
Where, $N$ is the total number of training samples, $C$ is the number of classes, 
and $h$, $w$, and $c$ denote the height, width, and number of channels of each image, respectively. We aim to prune a dense student network \(\theta_S\) under teacher supervision \(\theta_T\), such that the resulting sparse model maintains strong performance under a high sparsity budget. Unlike magnitude-based heuristics or post-hoc distillation, our approach integrates knowledge distillation (KD) directly into the pruning pipeline through teacher-informed importance score estimation. The teacher \(\theta_T\) is a high-capacity model pre-trained on the classification task, while the student \(\theta_S\), initialized with pre-trained weights, is first fine-tuned via KD with a combined objective of cross-entropy and Context-Aware Kullback-Leibler Divergence (CA-KLD).  

\begin{figure}[t]
    \centering
    % Increase width closer to full column and adjust trimming
    \includegraphics[width=\columnwidth, trim=21 21 19 22, clip]{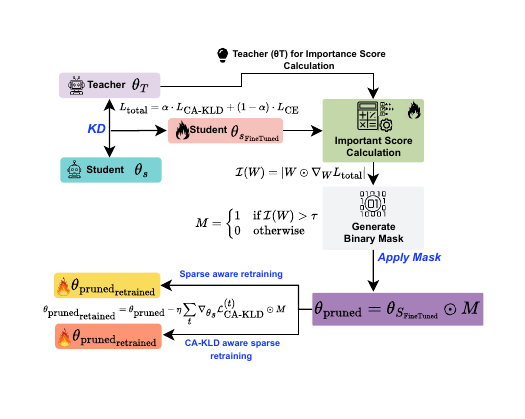}
    \caption{%
        Overview of the proposed teacher-guided one-shot pruning framework.
        The student is first trained with a combined KD and task loss, then a gradient-based importance score is used to generate a binary pruning mask. The pruned model is retrained under two regimes: (1) standard fine-tuning and (2) KD-guided retraining, to recover performance while maintaining sparsity.
    }
    \label{fig:pruning_framework}
\end{figure}

After fine-tuning, we compute a parameter-wise importance score for $\theta_S$ as
\begin{equation}
I_{\text{raw}} = \Psi(\theta_S, \theta_T; L),
\end{equation}
where $L$ is the total loss, $L_{\text{Total}}$, combining cross-entropy and CA-KLD, and $\Psi(\theta_S, \theta_T; L)$ is the importance score function.  
To stabilize against batch-wise noise, exponential moving average (EMA) smoothing with bias correction is applied (details in Subsection~\ref{subsec:teacher_guided_gradient}). The stabilized score at iteration $t$ is
\begin{equation}
I_{\text{final}} = \Phi_{\text{final}}(S; \gamma, t),
\end{equation}
where $\gamma \in [0, 1)$ is the decay factor, and $\Phi_{\text{final}}(S; \gamma, t)$ denotes the EMA with bias correction.
% \textcolor{orange}{Here, the dot operator ($\odot$) denotes element-wise (Hadamard) multiplication between the corresponding weight and its gradient.}
Using the aggregated scores, we construct a global pruning mask \(M \in \{0,1\}^{|\theta_S|}\) by retaining the top \((1-r)\times100\%\) weights, where $r$ is the sparsity target. The pruned model is then obtained as \(\theta_S^{\text{pruned}} = \theta_S \odot M\)(Here, the dot operator ($\odot$) denotes element-wise multiplication). Finally, retraining is performed (with and without KD) under the fixed mask to recover performance while strictly preserving sparsity. This enables one-shot pruning guided by both task and teacher signals, offering a more efficient and informed alternative to iterative pruning (see Fig.~\ref{fig:pruning_framework}).

\subsection{Knowledge Distillation Using CA-KLD Loss with Logit Normalization}
\label{subsec:cakld_based_kd}

We employ a knowledge distillation (KD) framework in which a compact student model is trained to mimic the behavior of a high-capacity teacher model. To enhance the effectiveness of distillation, we adopt the Context-Aware Kullback–Leibler Divergence (CA-KLD) loss~\cite{du2024bitdistillerunleashingpotentialsub4bit} in combination with logit normalization~\cite{sun2024logit}, ensuring the training signal remains both stable and informative. This same loss is later used to guide the pruning process through gradient-based importance estimation.

\subsubsection{Logit Normalization and Temperature Scaling}

Let $\mathbf{z}_T$ and $\mathbf{z}_S$ denote the teacher and student logits, respectively. Before applying softmax, we normalize these logits to reduce distributional variance and emphasize informative features:

\begin{equation}
\text{normalize}(\mathbf{z}) = \frac{\mathbf{z} - \mu_{\mathbf{z}}}{\sigma_{\mathbf{z}} + \epsilon}, \quad
\mu_{\mathbf{z}} = \mathbb{E}[\mathbf{z}], \quad 
\sigma_{\mathbf{z}} = \sqrt{\mathbb{E}[(\mathbf{z} - \mu_{\mathbf{z}})^2]}
\end{equation}

where $\epsilon$ ensures numerical stability. The normalized logits are then scaled by a temperature $T > 1$ to soften the probability distribution:

\begin{equation}
\mathbf{z}_{T}^{\tau} = \frac{\text{normalize}(\mathbf{z}_T)}{T}, \quad
\mathbf{z}_{S}^{\tau} = \frac{\text{normalize}(\mathbf{z}_S)}{T}
\end{equation}

\subsubsection{Context-Aware Distillation Loss}

Using the softened logits, we compute probability distributions via softmax:

\begin{equation}
\mathbf{P}_T = \text{softmax}(\mathbf{z}_{T}^{\tau}), \quad 
\mathbf{P}_S = \text{softmax}(\mathbf{z}_{S}^{\tau})
\end{equation}

The CA-KLD loss combines forward and reverse Kullback–Leibler divergences to ensure both models are aligned in terms of prediction confidence and uncertainty:

\begin{align}
\mathcal{L}_{\text{Fwd-KL}} &= \sum_{c=1}^C P_T(c) \log \frac{P_T(c)}{P_S(c)} \\
\mathcal{L}_{\text{Rev-KL}} &= \sum_{c=1}^C P_S(c) \log \frac{P_S(c)}{P_T(c)} \\
\mathcal{L}_{\text{CA-KLD}} &= \beta \cdot \mathcal{L}_{\text{Rev-KL}} + (1 - \beta) \cdot \mathcal{L}_{\text{Fwd-KL}}
\end{align}

To stabilize the gradient magnitude during training, the loss is scaled by the square of the temperature:

\begin{equation}
\mathcal{L}_{\text{CA-KLD}} \leftarrow \mathcal{L}_{\text{CA-KLD}} \cdot T^2
\end{equation}

\subsubsection{Training Objective}

The final training objective incorporates both the distillation signal and the supervised cross-entropy loss:

\begin{equation}
\mathcal{L}_{\text{Total}} = \alpha \cdot \mathcal{L}_{\text{CA-KLD}} + (1 - \alpha) \cdot \mathcal{L}_{\text{CE}}
\end{equation}

where $\alpha \in [0, 1]$ determines the balance between teacher-driven soft supervision and hard label learning. This combined objective serves not only to train the student effectively but also provides informative gradients for pruning via importance score computation.

\subsection{Teacher-Guided Gradient Importance Calculation}
\label{subsec:teacher_guided_gradient}

\begin{figure*}[ht]
  \centering
  \includegraphics[width=1.00\textwidth,height=0.2\textheight]{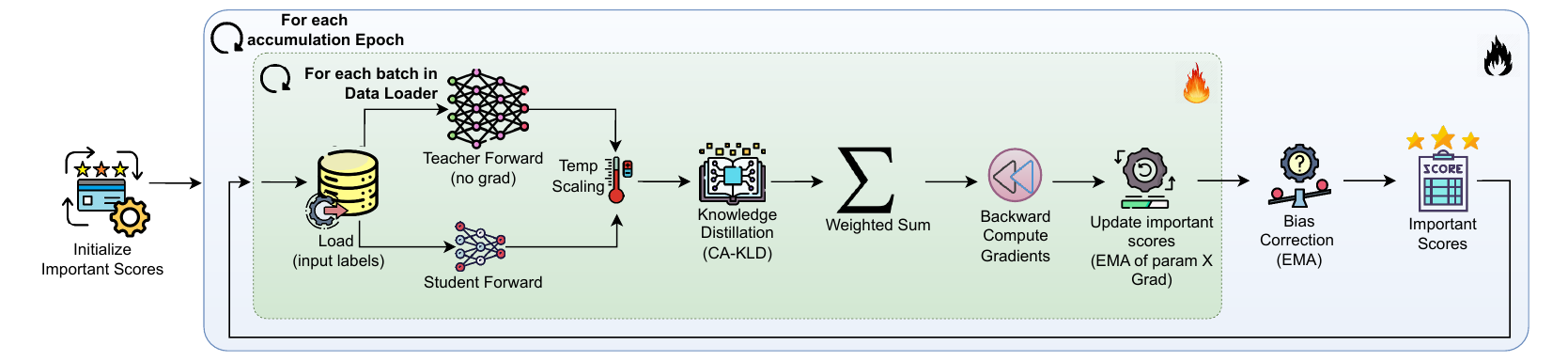} 
  \caption{Teacher Guided Important Score Computation }
  \label{fig:important_score}
\end{figure*}

This method identifies important weights for pruning by computing parameter importance scores based on the gradient flow induced by a joint training signal that combines both supervised and distillation objectives. As detailed in Section~\ref{subsec:cakld_based_kd}, the total loss $\mathcal{L}_{\text{Total}}$ incorporates both the cross-entropy loss and the CA-KLD distillation loss, allowing the teacher model to guide the student not only during learning but also in identifying the parameters most critical for task performance. The pruning decision is thus driven by the interaction between parameter magnitudes and their gradient alignment with respect to this joint loss, and the resulting raw importance scores are stabilized over time using an Exponential Moving Average (EMA) with bias correction. The overall process of computing gradient-based importance is depicted in Fig.~\ref{fig:important_score} and Algorithm~\ref{alg:importance_algo}.

To identify weights that are critical for both task performance and alignment with the teacher model, we compute teacher-guided gradient-based importance scores. As previously defined in Section~\ref{subsec:cakld_based_kd}, the total loss $\mathcal{L}_{\text{Total}}$ integrates both supervised and distillation signals, enabling the teacher to influence not only the student’s learning process but also the assessment of parameter relevance. This formulation allows the pruning process to retain weights that are essential for task success and effective teacher-student alignment.

The corresponding gradient with respect to the model parameters $W$ is then:
\begin{equation}
\nabla_W \mathcal{L}{\text{Total}} = \alpha \cdot \nabla_W \mathcal{L}{\text{CA-KLD}} + (1 - \alpha) \cdot \nabla_W \mathcal{L}_{\text{CE}},
\end{equation}
where $\alpha \in [0,1]$ balances the contributions of the distillation and supervised objectives. This gradient captures both the teacher’s guidance and the ground-truth supervision, allowing us to assess how essential each parameter is to the combined learning signal. This formulation corresponds to the function $\Psi(\theta_S, \theta_T; \mathcal{L})$ introduced in the problem formulation. To quantify the relevance of each weight, we compute the element-wise product between the weight and its gradient magnitude, yielding a raw importance score as introduced in the problem formulation:
\begin{equation}
I_{\text{raw}}(W) = \left| W \odot \nabla_W \mathcal{L}_{\text{Total}} \right|.
\end{equation}

Because raw gradient values are inherently noisy across batches, we smooth them using an exponential moving average (EMA):
\begin{equation}
I_t(W) = \gamma \cdot I_{t-1}(W) + (1 - \gamma) \cdot I_{\text{raw}}(W),
\end{equation}
where $\gamma = 0.9$ controls the rate of decay and determines how much past importance values influence the current estimate. To account for the initialization bias introduced by EMA in early steps, we apply a bias correction factor that normalizes the smoothed score:
\begin{equation}
I_{\text{final}}(W) = \frac{I_t(W)}{1 - \gamma^t},
\end{equation}
where $t$ is the number of batches seen. This bias-corrected EMA corresponds to $\Phi_{\text{final}}(S; \gamma, t)$ as defined in the problem formulation. The resulting score $I_{\text{final}}(W)$ serves as a robust, teacher-guided measure of parameter importance and forms the basis for global pruning in our pipeline.

\subsection{Global Pruning via Gradient Magnitude Thresholding}
\label{subsec:Thresholding}
Perform a one-shot, globally-aware pruning of the model using precomputed teacher-guided importance scores I(W) (Section \ref{subsec:teacher_guided_gradient}), enabling high sparsity while preserving performance-critical parameters.
% The process shown in and Algorithm \ref{alg:global_prune}
Unlike iterative pruning, which gradually removes weights and retrains over multiple cycles, our method applies aggressive, one-shot sparsification by pruning a portion (e.g., 95\%) of the model's weights in a single forward pass. The pruning decision is guided by gradient-based importance scores that encode both task supervision and teacher knowledge. While this one-shot approach offers efficiency and simplicity, it comes at the cost of an initial performance drop due to the severity of the pruning. To address this, the pruned model undergoes post-pruning retraining, allowing it to recover and adapt to its new sparse structure.

\subsubsection*{Step-by-Step Algorithm}

\subsubsection{Compute Global Threshold for Pruning}

Let \( v = \bigcup_{l} \text{vec}(I_l) \in \mathbb{R}^D \) denote the concatenated importance scores across all prunable weights in the network, where \( \text{vec}(\cdot) \) flattens the layer-wise importance tensor \( I_l \), and \( D \) is the total number of prunable parameters. To prune a target percentage \( p\% \) of the student model’s weights, compute the global pruning threshold \( \tau \) by identifying the 
\( k = \left\lfloor (1 - p) \cdot D \right\rfloor \)-th smallest value in \( v \): 
\(\tau = \text{TopK}\big(v,\, k = \lfloor (1 - p) \cdot D \rfloor\big)_{\min}\). 
This ensures that exactly \( p\% \) of the weights with the lowest importance scores are pruned globally.

\subsubsection{Generate Binary Pruning Mask}

The binary pruning mask \( M(W) \) is generated by thresholding the importance scores \( I(W) \), which quantify the contribution of each weight to both task performance and knowledge distillation fidelity. Given the computed global threshold \( \tau \), weights are retained if their importance exceeds \( \tau \), and pruned otherwise. Formally, for a weight \( W \) in layer \( l \), the binary mask is defined as

\begin{equation*} % use equation* for no extra space if numbering not needed
M_l(W) =
\begin{cases}
1 & \text{if } I_l(W) > \tau \quad (\text{retained}) \\
0 & \text{otherwise} \quad (\text{pruned})
\end{cases}
\end{equation*}

This enforces a global sparsity pattern while preserving the most critical weights, as identified by the teacher-guided gradient importance mechanism (see Section~\ref{subsec:teacher_guided_gradient}).

\subsubsection{Apply Mask to Prune the Model}
The final pruned student model $\theta_S^{\text{pruned}}$ is obtained by 
element-wise masking of the original weights: 
\(\theta_S^{\text{pruned}} = \theta_S \odot M\),
where $\odot$ denotes element-wise multiplication.All weights with importance scores below the global threshold are masked out (set to zero), resulting in a sparse model.

\begin{algorithm}[H]
\caption{Teacher Guided Gradient-Based Importance Score Calculation for Conv Layers}
\label{alg:importance_algo}
\textbf{Requires:} Teacher \(\theta_T\), Student \(\theta_S\), Dataset $\mathcal{D}$ , Temperature $\tau$, Distillation weight $\alpha$, Mixing factor $\beta_{\text{prob}}$, Epochs $E$, Logits $\hat{z}$ \\
\textbf{Output:} Layer-wise importance scores $I$
\begin{algorithmic}[1]
\State Initialize $I_l \gets 0$ for all conv layers $l$ in $\theta_{S}$, momentum $\mu \gets 0.9$, batch counter $b \gets 0$
\State Set $\theta_{T}$ to eval mode, $\theta_{S}$ to train mode.
\For{epoch $=1$ to $E$}
    \For{$(x, y) \in \mathcal{D}$}
        \State Move $(x, y)$ to Device, zero gradients in $\theta_{S}$
        \State \texttt{with torch.no\_grad():} $\hat{z}_T \gets \theta_{T}(x)$
        \State $\hat{z}_S \gets \theta_{S}(x)$
        \State Scale logits: $\hat{z}_T^\tau \gets \hat{z}_T / \tau$, $\hat{z}_S^\tau \gets \hat{z}_S / \tau$
        \State $L_{\text{KD}} \gets \text{CA-KLD}(\hat{z}_S^\tau, \hat{z}_T^\tau, \beta_{\text{prob}}) \cdot \tau^2$
        \State $L_{\text{CE}} \gets \text{CrossEntropy}(\hat{z}_S, y)$
        \State $L \gets \alpha \cdot L_{\text{KD}} + (1 - \alpha) \cdot L_{\text{CE}}$
        \State Backprop: $L.\texttt{backward}()$, $b \gets b + 1$
        \For{each conv layer $l$ in $\theta_{S}$}
            \If{$\nabla_l \neq \text{None}$}
                \State Compute layer contribution: $\Delta_l \gets |\theta_l \cdot \nabla_l|$
                \State (EMA): $I_l \gets \mu \cdot I_l  + (1 - \mu) \cdot \Delta_l$
            \EndIf
        \EndFor
    \EndFor
\EndFor
\For{each layer $l$} Normalization \State $I_l \gets I_l / (1 - \mu^b)$ \EndFor
\State \Return $I$
\end{algorithmic}
\end{algorithm}

\subsection{Sparse-Aware Retraining}

After pruning, the student model is retrained to restore accuracy while strictly maintaining its sparsity pattern. To achieve this, we apply two key mechanisms: masked gradient updates and momentum correction.

First, during backpropagation, gradients corresponding to pruned weights are masked out to prevent updates. For each layer \( l \), the masked gradient is computed as:
\begin{equation}
    \nabla W^{l}_{\text{masked}} = \nabla W^l \odot M^l
\end{equation}
where \( M^l \) is the binary pruning mask and \( \odot \) denotes element-wise multiplication. This ensures that no learning signal reaches the pruned weights.

Second, to avoid reactivation of pruned weights through residual momentum, we apply momentum correction. For SGD optimizers with momentum \( \mu \), the velocity update is constrained by the same binary mask:
\begin{equation}
    v^l(t+1) = \mu \cdot v^l(t) \odot M^l + \nabla W^{l}_{\text{masked}}
\end{equation}
This guarantees that momentum is only accumulated for unpruned parameters, preserving sparsity throughout retraining.

\subsubsection{KD-Aware Sparse Retraining}
\label{subsubsec:kd_aware}

We further enhance retraining through knowledge distillation, using the CA-KLD loss with logit normalization and temperature scaling to align the sparse student with the dense teacher. This process guides the pruned student to mimic teacher outputs while adhering to the fixed pruning mask. The student parameters are updated as:
\begin{equation}
    \theta_S^{\text{final}} = \theta_S^{\text{pruned}} - \eta \sum_t \nabla_{\theta_S} \mathcal{L}_{\text{CA-KLD}}(t) \odot M
\end{equation}
where \( \theta_S^{\text{pruned}} \) are the weights of the sparse student, \( \mathcal{L}_{\text{KD}} \) is the CA-KLD distillation loss, \( \eta \) is the learning rate, and \( M \) is the global pruning mask.

By enforcing sparsity constraints during optimization and coupling it with teacher-guided supervision, this retraining strategy enables the student to recover or even surpass pre-pruning accuracy without violating its compressed architecture.

\begin{table}[t]
    \centering
    \captionsetup{skip=5pt, labelsep=colon, justification=centering, singlelinecheck=false}
    \caption{Teacher and Student Model's Accuracies Across Different Datasets}
    \label{tab:table_one}

    \resizebox{\columnwidth}{!}{%
    \begin{tabular}{@{}l
                            S[table-format=2.2, table-alignment=center]
                            S[table-format=2.2, table-alignment=center]
                            S[table-format=2.2, table-alignment=center]@{}}
    \toprule
    Dataset & {ResNet50 (Teacher) (\%)} & {ResNet18 (Student) (\%)} & {ResNet34 (Student) (\%)} \\
    \midrule
    CIFAR-10       & 95.40 & 95.92 & 96.50 \\
    CIFAR-100      & 80.89 & 81.12 & 82.77 \\
    TinyImageNet   & 78.46 & 62.19 & {-} \\
    \bottomrule
    \end{tabular}%
    }
\end{table}

\section{Experiments and Results}
\label{sec:exp}

All experiments were conducted on a single NVIDIA RTX 3090 GPU (24 GB VRAM) with 16 vCPUs and 125 GB RAM. This setup provided sufficient computational resources to support both dense and sparse model training, ensuring reproducible results across pruning and distillation scenarios.

\subsection{Datasets}
We evaluate our approach on three widely used image classification benchmarks datasets:

\textbf{CIFAR-10 \cite{krizhevsky2009learning}:} CIFAR-10 contains 60{,}000 color images (32$\times$32) across 10 balanced classes, split into 50{,}000 training and 10{,}000 test samples. Its low resolution and limited categories make it suitable for lightweight model evaluation.

\textbf{CIFAR-100 \cite{krizhevsky2009learning}:} CIFAR-100 has the same size and resolution as CIFAR-10 but with 100 classes (600 images each). This finer granularity and fewer samples per class pose greater challenges in learning discriminative features under data scarcity.

\textbf{Tiny ImageNet \cite{le2015tiny}:} Tiny ImageNet includes 100{,}000 images (64$\times$64) over 200 classes, with 500 training samples each. Its higher resolution and class diversity approximate real-world recognition tasks while remaining computationally feasible, making it a common benchmark for efficient architectures.

\subsection{Training Settings}

\subsubsection{Gradient Importance Calculation Setting}

Gradient importance scores are computed with fixed hyperparameters for stability and reproducibility, where the balancing factor $\alpha = 0.7$ emphasizes the CA-KLD loss over cross-entropy, aligning student predictions with the teacher while maintaining task performance. The bidirectional KL in CA-KLD uses $\beta = 0.5$, giving equal weight to preserving teacher knowledge (forward KL) and reducing student overconfidence (reverse KL). Importance scores are accumulated over 3 epochs and smoothed using exponential moving averages (EMA, $\gamma = 0.9$) to reduce batch-level noise. For non-KD retraining, a temperature $T=5.0$ stabilizes gradients, while KD retraining uses two configurations: $(T=3.0, \alpha=0.7)$ and $(T=5.0, \alpha=0.7)$.These hyperparameter values ($\alpha$, $\beta$, and $\gamma$) were determined through rigorous hyperparameter tuning to ensure optimal stability and performance.

\subsubsection{Post-Pruning Training Setting}

To recover accuracy after aggressive pruning, pruned models are retrained with: (i) standard fine-tuning without KD using early stopping (patience=5 epochs), and (ii) KD-based fine-tuning using the total loss $L_{\text{Total}}$ combining task-specific loss and CA-KLD. Two KD configurations mirror the gradient importance settings. This approach ensures balanced supervision from ground-truth labels and teacher outputs, enhancing performance across all sparsity levels while maintaining stability and generalization.

% --- END: Corrected Merged Comparison Table Code ---59.29
\begin{table}[htbp]
    \centering
    \caption{Performance of Sparse Student (ResNet18-11.69M parameters) when Teacher (ResNet50) across CIFAR-10, CIFAR-100, and TinyImageNet (Our Method). 
    }
    \label{tab:table_two}
    \scriptsize
    \begin{tabular}{@{}
                    l
                    S[table-format=1.2, table-alignment=center] % Params
                    S[table-format=2.2, table-alignment=center] % Sparsity
                    S[table-format=2.2, table-alignment=center] % Accuracy
                    S[table-format=2.2, table-alignment=center] % Epoch
                    @{}}
    \toprule
    {\makecell{Dataset}} &
    {\makecell{Params \\ (M)}} &
    {\makecell{Sparsity \\ (\%)}} &
    {\makecell{Acc (\%)}} &
    {\makecell{\# Epoch \\ (Retrain)}} \\
    \midrule
    \multirow{6}{*}{CIFAR-10} 
        & 0.18 & 98.41 & 90.79 & 39 \\
        & 0.71 & 93.92 & 94.28 & 30 \\
        & 1.17 & 90.00 & 94.97 & 10 \\
        & 2.34 & 80.00 & 95.44 & 18 \\
        & 2.92 & 75.00 & 95.62 &  9 \\
        & 5.79 & 50.46 & 96.08 &  8 \\
    \midrule
    \multirow{6}{*}{CIFAR-100} 
        & 0.23 & 98.01 & 67.06 & 24 \\
        & 0.76 & 93.53 & 76.35 & 25 \\
        & 1.10 & 90.55 & 77.29 & 26 \\
        & 2.27 & 80.60 & 79.54 & 18 \\
        & 2.96 & 74.69 & 80.14 & 19 \\
        & 5.76 & 50.75 & 80.99 & 10 \\
    \midrule
    \multirow{6}{*}{TinyImageNet} 
        & 0.29 & 97.56 & 50.64 & 19 \\
        & 0.81 & 93.11 & 53.79 & 19 \\
        & 1.26 & 89.22 & 52.36 &  8 \\
        & 2.42 & 79.31 & 56.54 & 20 \\
        & 3.00 & 74.35 & 57.42 & 13 \\
        & 5.84 & 50.02 & 59.29 & 22 \\
    \bottomrule
    \end{tabular}
\end{table}

\section{Results}
\subsection{Teacher and Student Model Performance Before Pruning}

Table~\ref{tab:table_one} shows performance of the teacher (ResNet50) and student models (ResNet18, ResNet34) prior to pruning. All models were initialized with ImageNet-pretrained weights and fine-tuned on the target datasets, with students trained under KD supervision. On CIFAR-10, ResNet50 achieved $95.40\%$, while ResNet18 and ResNet34 slightly outperformed with $95.92\%$ and $96.50\%$, respectively. For CIFAR-100, ResNet50 reached $80.89\%$, compared to $81.12\%$ for ResNet18 and $82.71\%$ for ResNet34. On TinyImageNet, ResNet50 achieved $78.46\%$, while ResNet18 obtained $62.19\%$.

\subsection{Main Results}
\label{sec:main_results}

We evaluate the proposed KD-guided pruning framework, which employs the CA-KLD loss function with $T=3$ and $\alpha=0.7$, on CIFAR-10, CIFAR-100, and TinyImageNet. Using ResNet18 as the student distilled from a ResNet50 teacher, the method demonstrates strong robustness across varying sparsity levels, from moderate compression to extreme pruning, as summarized in Table~\ref{tab:table_two}. On CIFAR-10, the framework maintains high accuracy even under severe pruning, achieving $90.79\%$ at $98.41\%$ sparsity. Accuracy improves as sparsity decreases, reaching $94.97\%$ at $90.00\%$ and peaking at $96.08\%$ at $50.46\%$. This highlights KD’s role in mitigating the performance degradation typically observed with aggressive pruning. For CIFAR-100, the student shows resilience on this more challenging dataset, recording $66.32\%$ at $98.01\%$ sparsity. Accuracy increases steadily with reduced sparsity, attaining $81.01\%$ at $50.75\%$. These results indicate that the framework effectively preserves generalization in classification tasks. On TinyImageNet, the model achieves $50.64\%$ accuracy at $97.56\%$ sparsity, improving to $59.29\%$ at $50.02\%$. This demonstrates adaptability to larger-scale tasks while maintaining discriminative power. Overall, the results confirm that KD-guided pruning enables deep networks to achieve high compression without sacrificing accuracy.

% --- FIGURE 1: CIFAR-10 ---
\begin{figure*}[t!] % [t!] tries to place it at the top of a page
    \centering
    % Replace 'cifar10_accuracy_comparison.pdf' with your actual filename
    \includegraphics[width=\textwidth]{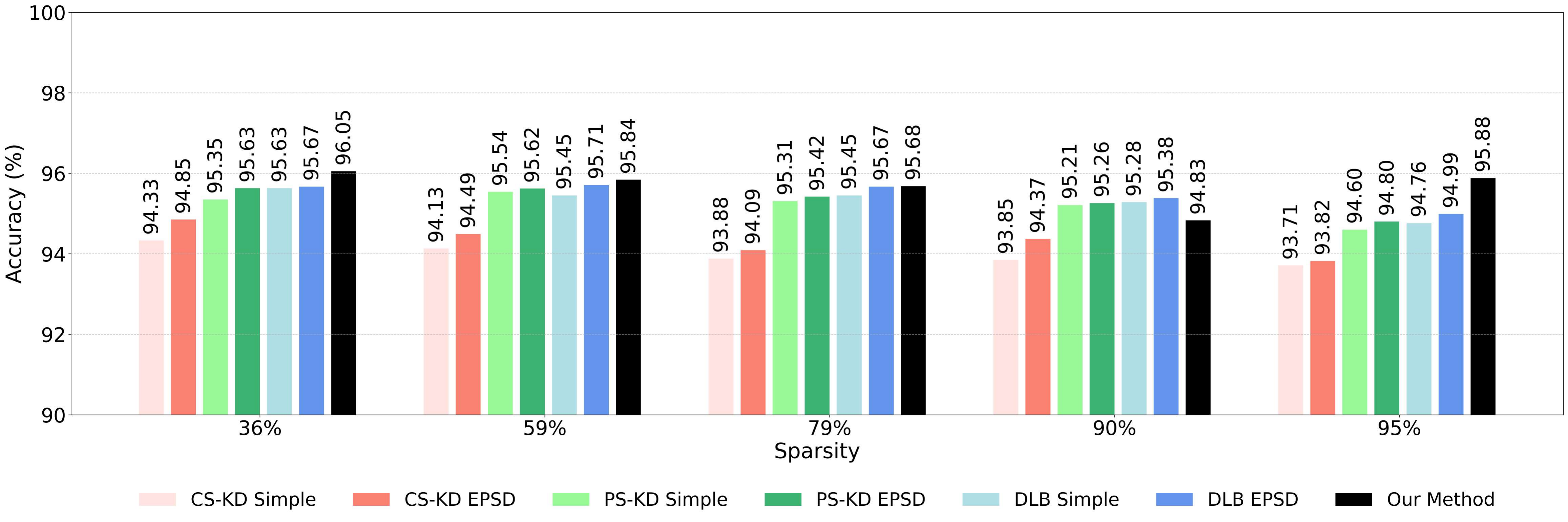}
   \caption{Top-1 accuracy of sparse ResNet-18 on CIFAR-10 at varying sparsity levels (36\% to 95\%), comparing our method against six baselines: CS-KD Simple~\cite{chen2024epsdearlypruningselfdistillation}~\cite{yun2020regularizing}, CS-KD EPSD~\cite{chen2024epsdearlypruningselfdistillation}, PS-KD Simple~\cite{chen2024epsdearlypruningselfdistillation}~\cite{kim2021paraphrasing}, PS-KD EPSD~\cite{chen2024epsdearlypruningselfdistillation}, DLB Simple~\cite{chen2024epsdearlypruningselfdistillation}~\cite{shen2022dynamic}, and DLB EPSD~\cite{chen2024epsdearlypruningselfdistillation}. Our method consistently outperforms all baselines at high and moderate sparsity levels.}

    \label{fig:cifar10_comparison}
\end{figure*}

% --- FIGURE 2: CIFAR-100 ---
\begin{figure*}[t!] % You might use [p] for page of floats if LaTeX struggles with placement
    \centering
    % Replace 'cifar100_accuracy_comparison.pdf' with your actual filename
    \includegraphics[width=\textwidth]{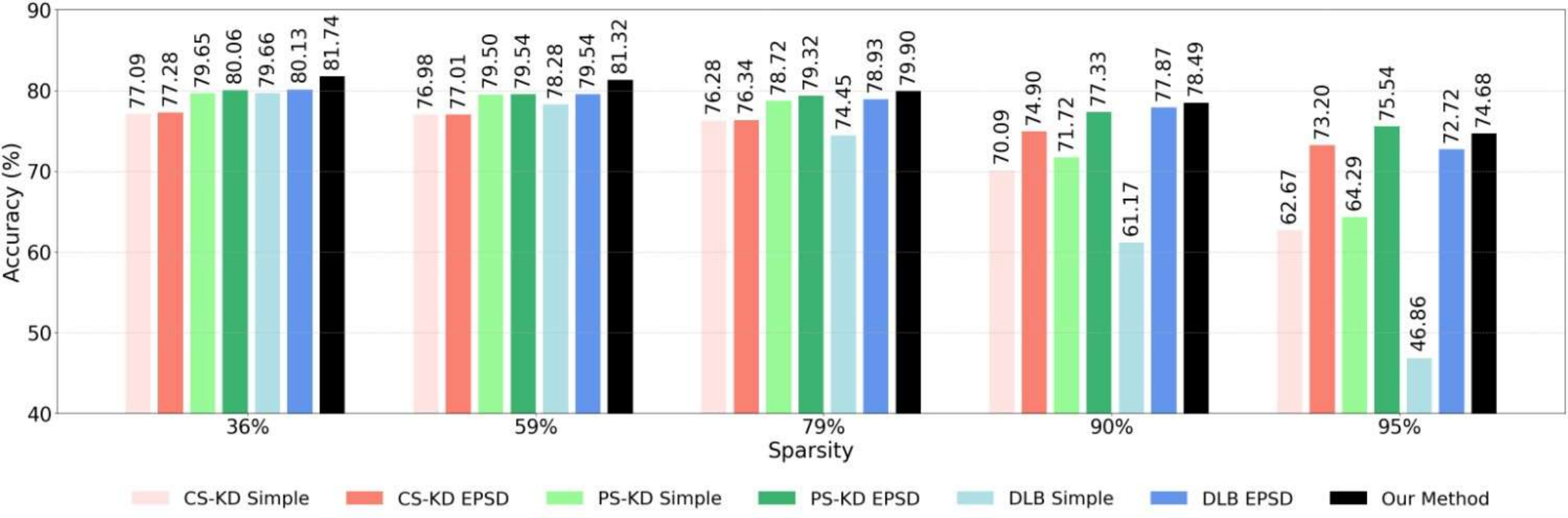}
  \caption{Top-1 accuracy comparison of ResNet-18 on CIFAR-100 across five sparsity levels (36\% to 95\%). Results are benchmarked against six baselines: CS-KD Simple~\cite{chen2024epsdearlypruningselfdistillation}~\cite{yun2020regularizing}, CS-KD EPSD~\cite{chen2024epsdearlypruningselfdistillation}, PS-KD Simple~\cite{chen2024epsdearlypruningselfdistillation}~\cite{kim2021paraphrasing}, PS-KD EPSD~\cite{chen2024epsdearlypruningselfdistillation}, DLB Simple~\cite{chen2024epsdearlypruningselfdistillation}~\cite{shen2022dynamic} , and DLB EPSD~\cite{chen2024epsdearlypruningselfdistillation}. Our method achieves consistently higher accuracy, especially at moderate sparsity levels.}

    \label{fig:cifar100_comparison}
\end{figure*}

% --- FIGURE 3: TinyImageNet ---
\begin{figure*}[t!]
    \centering
    % Replace 'tinyimagenet_accuracy_comparison.pdf' with your actual filename
    \includegraphics[width=\textwidth]{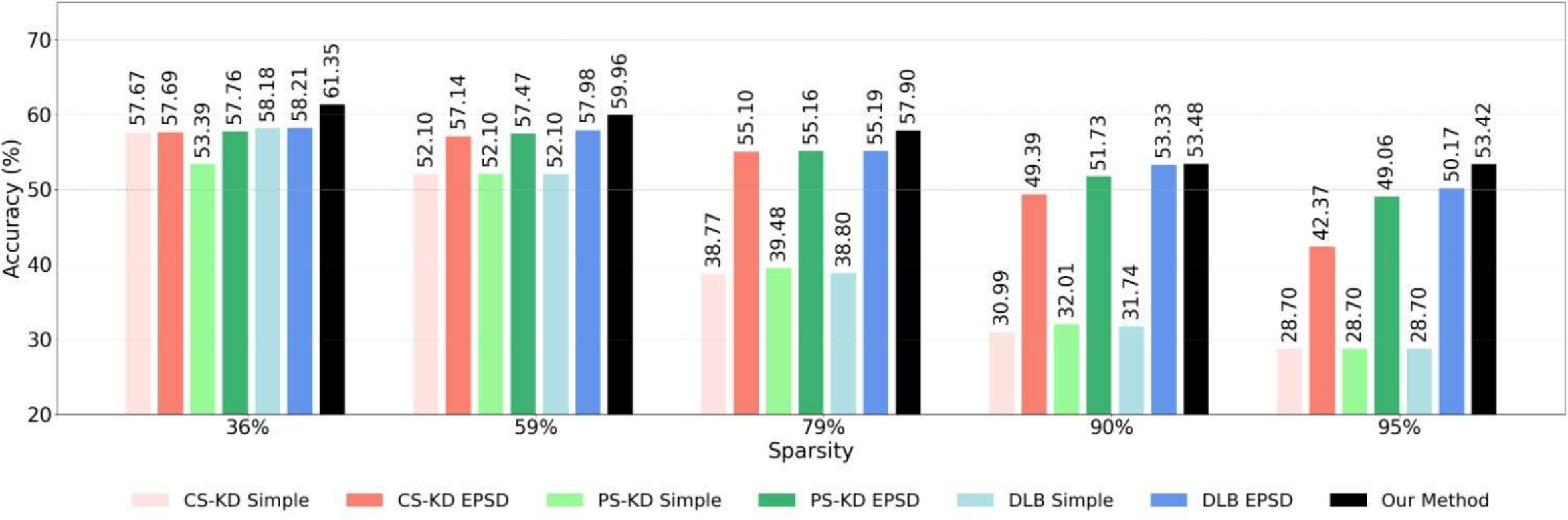}
\caption{Accuracy comparison of ResNet-18 on TinyImageNet across five sparsity levels. Our method consistently surpasses six distillation-based baselines: CS-KD Simple~\cite{chen2024epsdearlypruningselfdistillation}~\cite{yun2020regularizing}, CS-KD EPSD~\cite{chen2024epsdearlypruningselfdistillation}, PS-KD Simple~\cite{chen2024epsdearlypruningselfdistillation}~\cite{kim2021paraphrasing}, PS-KD EPSD~\cite{chen2024epsdearlypruningselfdistillation}, DLB Simple~\cite{chen2024epsdearlypruningselfdistillation}~\cite{shen2022dynamic}, and DLB EPSD~\cite{chen2024epsdearlypruningselfdistillation}. Even under high sparsity, our method maintains competitive accuracy, demonstrating effective knowledge preservation.}
    \label{fig:tinyimagenet_comparison}
\end{figure*}

\section{State-of-the-art Comparison}
\label{sec:comparison}
\subsection{Comparison With six Different Methods}

We evaluate our pruning-and-distillation framework against six different method baselines: CS-KD Simple \cite{yun2020regularizing} \cite{chen2024epsdearlypruningselfdistillation}, CS-KD EPSD \cite{chen2024epsdearlypruningselfdistillation}, PS-KD Simple \cite{kim2021paraphrasing} \cite{chen2024epsdearlypruningselfdistillation}, PS-KD EPSD \cite{chen2024epsdearlypruningselfdistillation}, DLB Simple \cite{shen2022dynamic} \cite{chen2024epsdearlypruningselfdistillation}, and DLB EPSD \cite{chen2024epsdearlypruningselfdistillation}. All experiments use ResNet-18 on CIFAR-100, TinyImageNet, and CIFAR-10, under five target sparsities. Results are displayed in Figures~\ref{fig:cifar10_comparison}, \ref{fig:cifar100_comparison}, and \ref{fig:tinyimagenet_comparison}.

\subsubsection{Comparison On CIFAR-100}
As shown in Figure~\ref{fig:cifar100_comparison}, our method outperforms all six competitive baselines across most sparsity levels. At 36\% sparsity, we achieve 81.74\% accuracy, exceeding the best baseline (DLB EPSD: 81.32\%) by +0.42 percentage points (pp). At 59\% sparsity, our accuracy remains strong at 81.32\%, outperforming all other methods including DLB EPSD (79.54\%, +1.78 pp). At 79\% sparsity, our method achieves 79.90\%, maintaining a slight lead over DLB EPSD (79.32\%) and PS-KD EPSD (78.72\%). Even under high sparsity (90\%), our method reaches 78.49\% accuracy, again outperforming DLB EPSD (77.87\%) and demonstrating resilience under compression. At the extreme 95\% sparsity level, our model retains 74.68\% accuracy—only slightly trailing PS-KD EPSD (75.54\%) by 0.86 pp, while significantly outperforming DLB EPSD (46.86\%) and all CS-KD variants.
\subsubsection{Comparison On TinyImageNet}
Figure~\ref{fig:tinyimagenet_comparison} shows the top-1 accuracy of ResNet18 across varying sparsity levels on Tiny ImageNet. At 36\% sparsity, our method achieves 61.35\%, outperforming the strongest baseline (DLB EPSD: 58.21\%) by +3.14 percentage points (pp). At 59\% sparsity, we retain 59.96\% accuracy again ahead of DLB EPSD (57.98\%) by +1.98 pp. As the pruning becomes more aggressive, our method maintains its advantage: at 79\% sparsity, we achieve 57.90\%, exceeding the closest baseline (PS-KD EPSD: 55.19\%) by +2.71 pp. Under extreme sparsity, our framework demonstrates notable resilience. At 90\% sparsity, we record 53.48\% accuracy, outperforming DLB EPSD (53.33\%) and all other baselines. At 95\% sparsity, our method reaches 53.42\%, a significant +3.25 pp improvement over DLB EPSD (50.17\%) and nearly +10 pp over the weakest-performing KD variants.

\subsubsection{Comparison On CIFAR-10}
As shown in Figure~\ref{fig:cifar10_comparison}, our method achieves top-1 accuracy across almost all sparsity levels when benchmarked against state-of-the-art pruning and distillation baselines, including CS-KD, PS-KD, and DLB (both simple and EPSD variants). At 36\% and 59\% sparsity, our method yields 96.05\% and 95.84\% accuracy, respectively, outperforming all baselines. Even under higher sparsity constraints, we retain leading performance: 95.68\% at 79\%, 94.83\% at 90\%, and 95.88\% at 95\% sparsity. Overall, across three benchmarks our approach achieves the best or highly competitive accuracy at low-to-moderate sparsities, and remains on par with state-of-the-art methods even under extreme pruning. This demonstrates the effectiveness of our gradient-based importance scoring and joint self-distillation strategy in preserving model performance under high compression.

\begin{table}[t] % Use table for single-column
    \centering
    \captionsetup{skip=5pt, labelsep=colon, justification=centering, singlelinecheck=false}
    \caption{Comparison of ResNet18 performance with COLT against Hossain et al.~\cite{hossain2022colt} across different datasets in terms of accuracy and latency.}
    \label{tab:table_three}
    \resizebox{\columnwidth}{!}{ % Automatically resize to column width
    \begin{tabular}{@{}l
                    S[table-format=2.2]
                    S[table-format=2.2]
                    S[table-format=3.2]
                    S[table-format=2.2]
                    S[table-format=2.2]
                    S[table-format=4.0]@{}}
    \toprule
    & \multicolumn{3}{c}{Ours (T=3/5, $\alpha$=0.7)} & \multicolumn{3}{c}{Hossain et al.~\cite{hossain2022colt}} \\
    \cmidrule(lr){2-4} \cmidrule(lr){5-7}
    Dataset      & {Sparsity.} & {Acc.} & {Latency.} & {Sparsity.} & {Acc.} & {Latency.} \\
                 & {(\%)}      & {(\%)} & {(min)}    & {(\%)}      & {(\%)} & {(min)}    \\
    \midrule
    CIFAR-10     & 97.7    & 91.87  & \textbf{27.82}  & 97.7    & \textbf{92.40} & 276 \\
    TinyImageNet & 97.4    & 51.14  & \textbf{42.43}  & 97.4    & \textbf{53.90} & 1756 \\
    CIFAR-100    & 97.4    & \textbf{68.66}  & \textbf{19.76}  & 97.4    & 68.40  & 355 \\
    \bottomrule
    \end{tabular}
    }
\end{table}

\subsection{Comparison with COLT}
\label{sec:comparison_colt}

To assess the effectiveness of our teacher-guided pruning framework at extreme sparsity levels, we benchmark it against Cyclic Overlapping Lottery Tickets (COLT) by Hossain et al.~\cite{hossain2022colt}, a state-of-the-art iterative pruning method that identifies overlapping sparse subnetworks through cyclic training and pruning phases. Table~\ref{tab:table_three} details the top-1 accuracy and computational latency for ResNet-18 on CIFAR-10, TinyImageNet, and CIFAR-100 datasets, comparing our best results against COLT-2 at matched sparsity ratios of 97.7\% for CIFAR-10 and 97.4\% for TinyImageNet and CIFAR-100. The empirical results reveal a compelling trade-off favoring our approach. While COLT-2 attains marginally higher accuracy on CIFAR-10 (92.40\% vs. our 91.87\%, a 0.53\% gap) and TinyImageNet (53.90\% vs. 51.14\%, a 2.76\% gap), our method surpasses it on CIFAR-100 (68.66\% vs. 68.40\%, a 0.26\% advantage). These accuracy differences are relatively minor, particularly considering the datasets' complexities, but the disparities in computational efficiency are stark. COLT-2's iterative cyclic process demands extensive training across multiple overlapping subnetworks, resulting in latencies of 276 minutes for CIFAR-10, 1756 minutes for TinyImageNet, and 335 minutes for CIFAR-100 factors of approximately 10$\times$, 41$\times$, and 18$\times$ higher than our respective latencies of 27.82, 42.43, and 19.76 minutes. For fair judgment, all latency measurements were conducted on an RTX 3090 GPU, consistent with the hardware used in COLT. This efficiency stems from our one-shot global pruning strategy, which leverages teacher-guided gradients (as detailed in Section~\ref{subsec:teacher_guided_gradient}) to perform a single, informed pruning step followed by concise retraining. In contrast, COLT's multi-phase iterations incur prohibitive overhead, limiting its scalability for rapid prototyping or deployment in time-sensitive scenarios. By achieving comparable or superior accuracy with drastically reduced computational demands, our framework demonstrates clear superiority, offering a more practical and resource-efficient solution for high-sparsity model compression in real-world, constrained environments.
\begin{table}[h!]
    \centering
    \caption{Comparison of Our Method and One-Shot LTH across different sparsity levels}
    \label{tab:table_four}
    \begin{tabular}{@{}l c c c@{}}
    \toprule
    Dataset & Sparsity (\%) & Our Method (Acc) & One-Shot LTH (Acc) \\
    \midrule
    \textit{CIFAR-10} & & & \\
     & 98.41 & \bfseries 90.79 & 89.47 \\
     & 93.92 & \bfseries 94.40 & 94.05 \\
     & 75.00 & \bfseries 95.62 & 94.85 \\
     & 50.00 & \bfseries 96.08 & 95.28 \\
    \midrule
    \textit{TinyImageNet} & & & \\
     & 98.41 & \bfseries 50.64 & 47.24 \\
     & 93.92 & 53.79 & \bfseries 55.43 \\
     & 74.35 & \bfseries 58.58 & 56.99 \\
     & 50.02 & \bfseries 60.93 & 56.91 \\
    \midrule
    \textit{CIFAR-100} & & & \\
     & 98.01 & \bfseries 67.06 & 65.20 \\
     & 80.60 & \bfseries 79.54 & 79.11 \\
     & 74.69 & \bfseries 80.14 & 79.63 \\
     & 50.63 & \bfseries 81.01 & 80.27 \\
    \bottomrule
    \end{tabular}
\end{table}

\subsection{Comparison with One-Shot Lottery Ticket Hypothesis}
\label{subsec:lth_comparison}

To further evaluate the efficacy of our teacher-guided pruning framework, we conduct a direct comparison against a one-shot variant of the Lottery Ticket Hypothesis (LTH)~\cite{frankle2018lottery}. In the one-shot LTH setting, the dense network is not fully pre-trained; instead, starting from the random initialization $\theta_0$, we first train for a short warm-up of $E_w$ epochs, then apply a single round of pruning to reach the target sparsity. The surviving weights are reset to their initial values $\theta_0$, and the resulting sparse subnetwork is trained to convergence. This baseline preserves the computational appeal of non-iterative pruning while aligning with the original LTH retraining protocol. Table~\ref{tab:table_four} reports the top-1 accuracy of our approach against the one-shot LTH baseline across different sparsity levels on CIFAR-10, CIFAR-100, and TinyImageNet using ResNet-18. Overall, our method provides consistent improvements over one-shot LTH, with the gap becoming more noticeable at higher sparsities. For example, at $98.41\%$ sparsity on CIFAR-10, our framework achieves $90.79\%$ accuracy compared to $89.47\%$ for LTH, a gain of $1.32\%$. Similar advantages are observed on TinyImageNet, where at $50\%$ sparsity we obtain $60.93\%$ accuracy versus $56.91\%$ for LTH, reflecting the benefit of teacher-guided importance estimation in more challenging datasets. Even at moderate sparsity levels (e.g., $50\%$), our method matches or slightly improves upon LTH, showing that the integration of KD signals into the pruning criterion provides stable performance gains without the need for iterative refinement. These results demonstrate that the use of teacher-informed gradients during the calculation of important scores leads to more reliable sparse models compared to other pruning methods. Additional, comparison with entropy-guided pruning (EGP) approach by Liao et al. \cite{liao2023can} given in the Appendix\footnote{Additional results: \url{https://github.com/sami0055/TGOSP-CAKD/tree/main}}

\section{Ablation Study}

To gain deeper insights into the effectiveness of knowledge distillation in the pruning setting, we compare student performance both without KD and with KD at two temperatures ($T=3$ and $T=5$). 
Table~\ref{tab:table_five} summarizes results for ResNet-18 students across CIFAR-10, CIFAR-100, and TinyImageNet under different sparsity levels. As discussed earlier, pruning alone (w/o KD) causes a significant loss in accuracy, particularly for high sparsity ratios and more complex datasets. For example, CIFAR-10 accuracy falls to $86.9\%$ at $98\%$ sparsity, while CIFAR-100 and TinyImageNet degrade to $62.5\%$ and $47.7\%$, respectively. Introducing KD clearly alleviates this issue. Across all datasets, KD consistently boosts performance, even at extreme sparsity. On CIFAR-10, KD recovers almost $+4\%$ at $98\%$ sparsity, raising accuracy from $86.9\%$ to $90.8\%$. Similar improvements are observed on CIFAR-100 (from $62.5\%$ to $67.1\%$) and TinyImageNet (from $47.7\%$ to $50.6\%$). At moderate sparsity (e.g., $75\%$--$90\%$), KD nearly closes the gap with dense models, demonstrating its robustness. 
Comparing temperatures, we find that both $T=3$ and $T=5$ provide substantial benefits, with $T=3$ performing slightly better on CIFAR-10 and CIFAR-100, while $T=5$ yields marginal gains on TinyImageNet. This suggests that the optimal temperature may depend on dataset complexity, but in all cases, KD stabilizes training and preserves accuracy under high compression. Overall, these results validate that while pruning alone is insufficient, our context-aware KD strategy enables highly sparse student models to remain competitive with their dense counterparts.

\section{Discussion}
Our one-shot global pruning pipeline, leveraging a teacher-guided CA-KLD signal and smoothed gradient-based importance scores, maintains high accuracy even at extreme sparsity (>98\%). For instance, on CIFAR-10 at 98.41\% sparsity, we achieve 90.79\% accuracy, surpassing EGP, which only outperforms us above 98\% sparsity due to its entropy criterion preserving critical neurons. We outperform six KD-based baselines (CS-KD Simple \& EPSD, PS-KD Simple \& EPSD, DLB Simple \& EPSD) by introducing KD in important score calculation, unlike their post-pruning KD application. Compared to iterative methods like COLT, our approach delivers comparable accuracy with significantly lower latency (e.g., 27.82 vs. 276 minutes on CIFAR-10 at 97.7\% sparsity). Against one-shot LTH, our superior importance scoring yields better results. Slight sensitivity to KD temperature (T=3 often outperforms T=5) and minor accuracy dips at extreme sparsity suggest exploring hybrid entropy-gradient criteria or adaptive-T schedules. Our framework effectively balances compression, accuracy, and efficiency, making it ideal for resource-constrained edge deployments.

\section{Conclusion}
\label{sec:conclusion}
We presented a teacher-guided one-shot pruning framework that integrates knowledge distillation directly into pruning through a gradient-based importance metric. By combining cross-entropy and context-aware KL divergence with logit normalization, we derive stable importance scores that enable aggressive and effective global pruning.  Across CIFAR-10, CIFAR-100, and TinyImageNet, our method preserves accuracy even under extreme sparsity and surpasses state-of-the-art pruning and distillation baselines. Compared to iterative pruning, it achieves competitive results at a fraction of the cost, offering a practical and scalable solution for resource-constrained environments. This work establishes the value of KD into pruning itself and opens pathways for extending the framework to broader architectures and tasks.

\begin{table}[t]
    \centering
    \caption{Ablation study: Accuracy (\%) of ResNet-18 students without KD and with KD ($T=3,5$) across sparsity levels.}
    \label{tab:table_five}
    \scriptsize
    \resizebox{\linewidth}{!}{%
    \begin{tabular}{l S[table-format=2.2] S[table-format=2.2] S[table-format=2.2] S[table-format=2.2]}
    \toprule
    {Dataset} & {Sparsity (\%)} & {Acc. w/o KD (\%)} & {Acc. w/ KD (T=3) (\%)} & {Acc. w/ KD (T=5) (\%)} \\
    \midrule
    \rowcolor{gray!10}
    \multirow{6}{*}{CIFAR-10} 
        & 98.41 & 86.99 & \textbf{90.79} & 90.66 \\
        & 93.92 & 92.97 & 94.28 & \textbf{94.40} \\
        & 90.00 & 94.12 & \textbf{94.97} & 94.83 \\
        & 80.00 & 95.21 & \textbf{95.44} & 95.29 \\
        & 75.00 & 95.46 & \textbf{95.62} & 95.60 \\
        & 50.46 & 96.06 & \textbf{96.08} & 96.03 \\
    \midrule
    \rowcolor{gray!10}
    \multirow{6}{*}{TinyImgNet} 
        & 97.56 & 47.71 & \textbf{50.64} & 50.07 \\
        & 93.11 & 51.74 & \textbf{53.79} & 52.11 \\
        & 89.22 & 52.53 & \textbf{52.36} & 52.19 \\
        & 79.31 & 53.41 & 56.54 & \textbf{57.95} \\
        & 74.35 & 54.39 & 57.42 & \textbf{58.58} \\
        & 50.02 & 58.53 & 59.29 & \textbf{60.93} \\
    \midrule
    \rowcolor{gray!10}
    \multirow{6}{*}{CIFAR-100} 
        & 98.01 & 62.49 & 66.32 & \textbf{67.06} \\
        & 93.53 & 72.00 & 75.34 & \textbf{76.35} \\
        & 90.55 & 73.94 & \textbf{77.40} & 77.29 \\
        & 80.60 & 77.61 & 79.44 & \textbf{79.54} \\
        & 74.69 & 78.95 & 79.78 & \textbf{80.14} \\
        & 50.75 & 80.69 & \textbf{81.01} & 80.99 \\
    \bottomrule
    \end{tabular}%
    }
\end{table}

% *** REFERENCES ***
% Use BibTeX for bibliography management
\bibliographystyle{IEEEtran}
\bibliography{ref} % Ensure this matches your .bib file name (e.g., ref.bib)
\end{document}